\pdfoutput=1
\relax

\documentclass[11pt]{article}

\usepackage[]{acl2023}
\usepackage[T1]{fontenc}
\usepackage{times}
\usepackage{latexsym}
\usepackage{bm,amsmath}

\usepackage[T1]{fontenc}

\usepackage[utf8]{inputenc}

\usepackage{microtype}

\usepackage{inconsolata}

\usepackage{algorithm}
\usepackage{algorithmicx}
\usepackage{algpseudocode}
\usepackage{booktabs}
\let\oldReturn\Return
\renewcommand{\Return}{\State\oldReturn}

\algdef{SE}
[OBJECT]
{Object}
{EndObject}
[1]
{\textbf{object} \textsc{#1}}
{\textbf{end object}}

\algnewcommand\algorithmicinput{\textbf{Input:}}
\algnewcommand\algorithmicoutput{\textbf{Output:}}
\algnewcommand\algorithmicparameter{\textbf{Parameters:}}
\algnewcommand\INPUT{\item[\algorithmicinput]}
\algnewcommand\OUTPUT{\item[\algorithmicoutput]}
\algnewcommand\PARAMETER{\item[\algorithmicparameter]}
\algrenewcommand\algorithmicindent{1.0em}
\usepackage[utf8]{inputenc}

\usepackage{microtype}
\usepackage{todonotes}
\usepackage{inconsolata}

\def\url#1{\expandafter\string #1}

\usepackage{hyperref} 

\author{
Vasudha Varadarajan$^1$,
Sverker Sikström$^2$,
Oscar N.E. Kjell$^2$,
H. Andrew Schwartz$^1$ \\
    $^1$Department of Computer Science, Stony Brook University \\
$^2$Department of Psychology, Lund University \\
{\tt \{vvaradarajan,has\}@cs.stonybrook.edu} \\
}

\title{ALBA: Adaptive Language-Based Assessments for Mental Health }

\begin{document}

\maketitle
\begin{abstract}

Mental health issues differ widely among individuals, with varied signs and symptoms. 
Recently, language-based assessments have shown promise in capturing this diversity, but they require a substantial sample of words per person for accuracy.
This work introduces the task of Adaptive Language-Based Assessment (\texttt{ALBA}), which involves adaptively \textit{ordering} questions while also \textit{scoring} 
an individual's latent psychological trait using limited language responses to previous questions.
To this end, we develop adaptive testing methods under two psychometric measurement theories:
\textit{Classical Test Theory} and \textit{Item Response Theory}. 
We empirically evaluate ordering and scoring strategies, organizing into two new methods: a semi-supervised item response theory-based method (\texttt{ALIRT}) and a supervised \textit{Actor-Critic} model. 
While we found both methods to improve over non-adaptive baselines, We found \texttt{ALIRT} to be the most accurate and scalable, achieving the highest accuracy with fewer questions (e.g., Pearson r $\approx$ 0.93 after only 3 questions as compared to typically needing at least 7 questions). 
In general, adaptive language-based assessments of depression and anxiety were able to utilize a smaller sample of language without compromising validity or large computational costs.

\end{abstract}

\section{Introduction}


 \begin{figure}[!t]
    \centering
\includegraphics[width=\columnwidth]{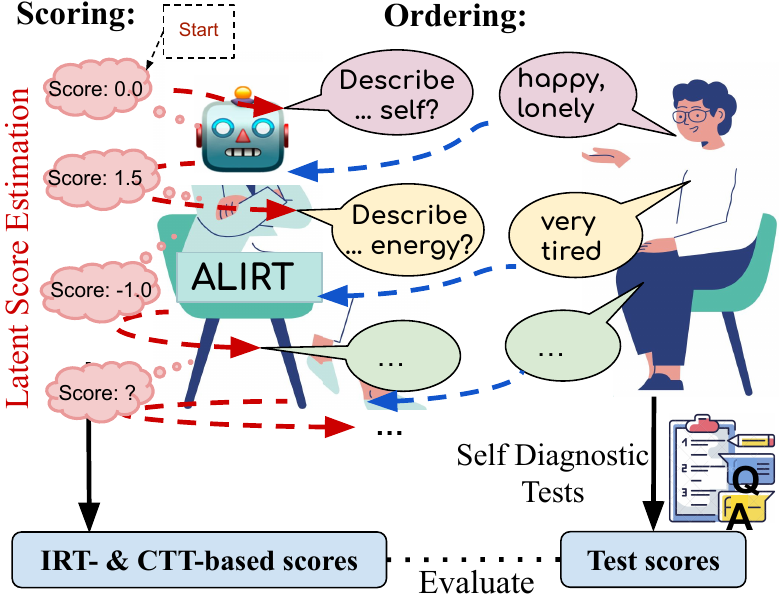} 
    \caption{
    The \texttt{ALBA} task: the system picks the most informative question to ask based on previous responses, much like a therapist would in real life. To do this, we introduce an IRT-based semi-supervised method, \texttt{ALIRT} and an \textit{Actor-Critic} model, and compare their performance with a limited set of language-response questions against self-report diagnostic questionnaire scores for depression and anxiety test scores (PHQ-9 and GAD-7).
    } 
    \label{fig:method_illusv1}
\end{figure}


Standard mental health (e.g., depression, anxiety) assessment consists of asking patients a fixed set of questions to which they respond along a rating scale. 
For example, the Patient Health Questionnaire (PHQ) asks, ``\textit{Over the last 2 weeks, how often have you been bothered by having little interest or pleasure in doing things?} 
0: Not at all, 1: Several days, 2: More than half the days, or 3: 
Nearly every day'' \cite{kroenke2001phq, siwek2009mood}. 
This presents an information limitation: answering a fixed set of questions often leads to some unnecessary questions, while logging answers on a single dimensional rating scale limits the total information content possible. 

Recent work has begun to address this information limitation by utilizing a patient's natural language to assess mental health conditions~\cite{milne2016clpsych,de2016discovering,eichstaedt2018facebook,kjell2019semantic}. 
Such open-ended language responses enable participants to elaborate on their mental health, where appropriate computational methods can be used to quantify the language response to a semantic scale \cite{kjell2019semantic}. 
Patients find that language is more precise in communicating their mental health issues, preferring it to rating scales~\cite{sikstrom2023precise}, but
 language-based assessments can be lengthy \cite{sikstrom2023precise} often requiring word minimums~\cite{eichstaedt2021closed}.
Therefore, not only is there merit in mapping patients' language
to their conditions but also in asking questions or prompting for language more optimally to reduce redundant information and adapt iteratively to each question.




We introduce the task of Adaptive Language-Based Assessment (\texttt{ALBA}). 
\texttt{ALBA} is inspired by adaptive testing used in psychology and education \cite{xu2020development}, which usually utilizes a Bayesian statistical framework known as Item Response Theory (IRT) over discrete-valued responses to questions.  It helps adapt to the future
prompts based on the prompts already administered
to the participant, iteratively picking the next best question to ask based on the current latent estimate.

 We envision \texttt{ALBA} as a step towards 
 the development of conversational diagnostic agents, as it allows for the modeling of language prompts in a semi-supervised manner. Using this approach, agents can adaptively conduct language-based assessments with dynamic scoring and benefit from a prior understanding of responses, leading to improved diagnostic accuracy, less use of patient and clinician time, and more personalized interactions.

Our main \textbf{contributions} include:
(1) introducing the task of adaptive language-based assessment (\texttt{ALBA}) for selecting questions that provide the most informative responses; 
(2) development of \texttt{ALIRT} model, an approach integrating predictive modeling to be able to apply IRT to linguistic responses (as opposed to numeric responses as is typically used) -- produces depression scores from only 4 question-responses that have 90\% of the variance explained of an assessment that uses 11 question-responses; \footnote{The code for \texttt{ALIRT} and other methods described in the paper can be found \href{https://github.com/humanlab/alba-irt}{here.}}
(3) development of an \textit{Actor-Critic} model model for adaptive language-based assessments; (4) evaluation of different modeling strategies within these models (e.g., discretizing in 2-tomous); (5)
extensive empirical comparison
of these methods with more straight-forward approaches covering multiple scoring strategies, and
(6) insights into the questions that generally produce responses with the most information (i.e. ability to distinguish participant depression severity)
informing better question/prompt creation.



\section{Background}
Item Response Theory estimates a latent variable as a proxy for an unobservable attribute (such as depression) by modeling the interaction between: (1) a latent variable (e.g. depression ``score''), (2) observable variables from a population of participants (e.g. responses to questions \footnote{A \textit{question} corresponds to an item in the Item Response Theory literature; therefore, the words ``item'' and ``question'' are used interchangeably in this paper. }  about lack of sleep, appetite, etc), and (3) the data points from a particular individual who is to be assessed. The latent score is typically estimated with Bayes parameter estimation.  
IRT addresses a short-coming of Classical Test Theory (CTT) \cite{lord2008statistical} which is based on the assumption that there is a \textit{true} score for an unobservable attribute, which is typically taken to be the summed value of all observable variables (e.g., ratings), and that the observed estimate is only
off of the true score by some error from the act of measurement. 
This ignores correlations between the observable variables. 
In this work, we compare scores based on both IRT and CTT while evaluating the proposed adaptive testing methods over standard questionnaires for measuring major depression (PHQ-9) and generalized anxiety disorder (GAD-7).\footnote{While the PHQ-9 and GAD-7 are still based on CTT, they are longer form questionnaires that check across symptoms described in the diagnostic and statistical manual, version 5~\cite{american2013diagnostic}.}



Various IRT modeling functions have been proposed for capturing discrete responses \cite{samejima2016graded, muraki2016generalized, chalmers2012mirt}. Increased model complexity in IRT leads to a rise in parameters, requiring larger datasets. Given the multidimensional nature of language representation and limited advances in simultaneous modeling across dimensions, we employ \textit{polytomous} item response theory \cite{ostini2006polytomous} to discretize language responses onto a graded scale.

 In this work, we introduce Adaptive Language-based IRT (\texttt{ALIRT}), which uses a supervised approach to polytomize the linguistic responses, and then employs adaptive testing using IRT.

\section{Methods}
The components that we develop for \texttt{ALBA} aims at  (1) dynamically ordering a set of questions, picking next-best at a time;  and (2) scoring the assessment at each step. 
These components can either be modeled jointly (as in \texttt{ALIRT}) or step-wise (as in our \textit{Actor-Critic} model). 
We describe both approaches below and a suite of more straightforward baseline approaches to compare against. 


\algrenewcommand\algorithmicindent{0.5em}
\begin{algorithm}[h!]

\caption{\texttt{ALIRT}}
\label{alg:method}
\small
\begin{algorithmic}[1]
\item[\textbf{Notation:}]
$J$: Num. of questions; $\texttt{K}$: Levels of rating scale; 
\item[\hspace{1em}]\texttt{poly}: polytomization split;
\texttt{tr}: train split; \texttt{te}: test split; 
\item[\hspace{1em}]M$^{\texttt{j}}$: Model takes in item j embeddings, fit to ``true'' values; 
\item[\hspace{1em}]D$_{\texttt{i}}^{\texttt{j}}$: data split i; word embedding of response to item j.

\item[\hspace{1em}]Y$_{\texttt{i}}$: Measure values (e.g. PHQ-9) for data split i.
\item[\hspace{1em}]$\hat{Y}^{j}_{\texttt{i}}$: Predicted measures for data split i based on item j.
\item[\hspace{1em}]O$_{\texttt{i}}^{\texttt{j}}$: data split i; polytomized response to item j. 
\item[\hspace{1em}] $\beta_{j}, \alpha$: Characteristic curve parameters for item j.

\vspace{1mm}
\hrule
\vspace{1mm}
\noindent \textbf{\textsc{Phase 1: Polytomization}}
\vspace{1mm}
\hrule
\vspace{1mm}


\Function{Polytomization}{Train set: \{D$_{\texttt{poly}}$, Y$_{\texttt{poly}}$\}, IRT train to polytomize:D$_{\texttt{tr}}$,  IRT test to polytomize:D$_{\texttt{te}}$}

  \For{$\texttt{j}=1,...,J$} 
\State M$^{\texttt{j}} \leftarrow $ FitRegression( D$_{\texttt{poly}}$, Y$_{\texttt{poly}})$
\State $\Hat Y_{\texttt{poly}} \leftarrow $ M$^{\texttt{j}}$.predict(D$_{\texttt{poly}})$
\State$ \Hat Y^{\texttt{j}}_{\texttt{tr}}  \leftarrow $ M$^{\texttt{j}}$.predict(D$_{\texttt{tr}})$
; $\Hat Y^{\texttt{j}}_{\texttt{te}}  \leftarrow 
 $  M$^{\texttt{j}}$.predict(D$_{\texttt{te}})$
\State thresholds $\leftarrow [(\frac{\texttt{100k}}{\texttt{K}}$)-percentile($\Hat Y_{\texttt{poly}});\forall \texttt{k}\in$\{1,2...$\texttt{K}$\}]
\State O$^{\texttt{j}}_{\texttt{tr}}\leftarrow$  min($\texttt{k}$) | thresholds[$\texttt{k}$] > $\Hat Y^{\texttt{j}}_{\texttt{tr}}$ (likewise for $\Hat Y^{\texttt{j}}_{\texttt{te}}$)
\EndFor
\Return O$_{\texttt{tr}}$, O$_{\texttt{te}}$

\EndFunction


\vspace{1mm}
\hrule
\vspace{1mm}
\noindent \textbf{\textsc{Phase 2: Adaptive Testing}}
\vspace{1mm}
\hrule
\vspace{1mm}
\State  $\sigma_{\textsc{IRT}}=(\beta_{1},\beta_{2},...\beta_{J}, \alpha $)$\leftarrow$ExpectationMaximization(O$_{\texttt{tr}}$)
\Function{GetNextQuestion}{$\theta$, itemsLeft}

\Return j | max. FisherInfo($\sigma^{\texttt{j}}_{\texttt{IRT}}$($\theta$)) $ \forall \texttt{j}\in$\{itemsLeft\}

\EndFunction
\Function{Update}{Item: j, Response: o$^{\texttt{j}}_{\texttt{te}}$} 
\State $\theta \leftarrow$ Maximum a Priori: maximize(P(o$^{\texttt{j}}_{\texttt{te}}$ ,$\sigma_{\texttt{IRT}}$ | $\theta$) P($\theta$))
\textbf{return} $\theta$
\EndFunction

\For {each set of user responses o$_{\texttt{te}}$ in O$_{\texttt{te}}$:}
\State $\theta$ = $\theta_{0}\leftarrow 0$; itemsLeft $\leftarrow$ \{1... J\} 
\While{itemsLeft} 
\State j $\leftarrow$ \textsc{GetNextQuestion}($\theta$, itemsLeft)
\State $\theta$ $\leftarrow $ \textsc{Update}( j, o$^{\texttt{j}}_{\texttt{te}}$);  itemsLeft $\leftarrow$ itemsLeft - \{j\}; 
\EndWhile
\EndFor



  

\end{algorithmic}
\label{alg: alirt-workflow}
\end{algorithm}

\subsection{Adaptive Language-based IRT (\texttt{ALIRT})} Adaptive Language-based IRT uses adaptive testing on language responses that are polytomized with a supervised model trained on word embeddings.
The process is implemented in three phases:

\paragraph{Polytomization} Language responses are multi-dimensional and can be represented as word embedding vectors. As discussed, we use \textit{polytomous} item response theory to discretize the language responses to a graded scale. The responses are polytomized by training supervised models for each item on one split of the dataset: $D_{\texttt{poly}}$.


Word embeddings are extracted for each question and response in $D_{\texttt{poly}}$. For each question, the participants in our dataset (\S\ref{sec:dataset}) are prompted for descriptive, context-independent words. Since contextual models 
aren't trained to represent individual descriptive words, we train our own word embeddings based on Principal Component Analysis over a term-document matrix with log-entropy weighting (aka Latent Semantic Analysis or LSA), allowing flexibility in choosing dimensions to represent language effectively. This approach has been proven as effective as word2vec, particularly in the context of psychology \cite{altszyler2016comparative}. The reduced dimensional space was 300, and the first 10 dimensions were used for the embeddings. The word embeddings were trained on a large dataset that contained similar word responses (69864 responses with 6728 unique words) to mental health questions-- the dataset and word representations are introduced in \cite{kjell2019semantic}. 
Preferring smaller embedding sizes which are suitable for low-resource domains like mental health, we utilize 10 dimensions for each question. For replicability, any comparable word embeddings could be used: we explore dimension-reduction on GloVe and RoBERTA-large embeddings as well in Appendix \ref{appendix: roberta_large}.



For each of the J questions, a multiple-ridge regression model is trained to predict the psychometric measure (PHQ-9 and GAD-7) on the averaged word embeddings of its responses. The average RMSE over all the ridge regression models is 10.93 for PHQ-9 and 8.64 for GAD-7. Each of the question's models is applied to the test set to predict the psychometric measures per question per sample. This is the \textit{supervised} aspect of our approach.

The predicted psychometric measures on $D_{\texttt{poly}}$ are thresholded based on percentiles given the discretization we wish to obtain for the questions. For example, if we want each question's responses to be polytomized from a scale of 1 to 3, i.e. [0,1,2,3], then the percentile thresholds are the quartiles for the predictions of the regression models for each question for $D_{\texttt{poly}}$. These thresholds are applied to the rest of the dataset, which is split into two more parts: $D_{\texttt{tr}}$ for training the adaptive testing model, and $D_{\texttt{te}}$ for evaluating the adaptive testing model.

 

\paragraph{Adaptive Testing with IRT} can be directly applied to the polytomized data. In terms of Item Response Theory, an \textit{item} is the question, and the corresponding \textit{response} is the polytomized language response. To train the model, all the item parameters are simultaneously fit on $D_{\texttt{tr}}$ using Broyden–Fletcher–Goldfarb–Shanno (BFGS) optimization algorithm \cite{liu1989limited} as 2PL unidimensional IRT \cite{lord2012applications} until convergence (See Appendix \ref{appendix:irt}). We utilize a well-known R package for IRT, {\texttt{mirt}}\footnote{\hyperlink{https://CRAN.R-project.org/package=mirt}{https://CRAN.R-project.org/package=mirt}}. For each data point in $D_{\texttt{te}}$, the testing is done by sequentially estimating the latent IRT variable for each question while keeping the learned IRT item parameters fixed. The latent variable is initialized at the average latent score of $D_{\texttt{tr}}$. To pick the next best question we utilize Fisher Information, a common criteria known to work well over most scenarios \cite{chalmers2016generating}. The latent variable estimate ($\hat L$) is an \textit{unsupervised} estimated value of the factor representing the selected questions.  

To make the method comparable to other strategies based on the Classical Test Theory, we also calculate the average of the predicted measures ($\widehat{Y}$). For each question that is picked iteratively by the adaptive testing algorithm, to predict in the same scale as the psychometric measures. 

We utilize \texttt{mirtCAT}\footnote{\hyperlink{https://CRAN.R-project.org/package=mirtCAT}{https://CRAN.R-project.org/package=mirtCAT}}, a computerized adaptive testing framework based on \texttt{mirt} to implement adaptive testing.

We run a 9-fold cross validation ($D_{\texttt{poly}}$:4, $D_{\texttt{tr}}$:4, $D_{\texttt{te}}$:1) across the two phases as described in the Algorithm \ref{alg: alirt-workflow}-- hence our approach is semi-supervised. Since the latent variable and the psychometric measures are on different scales, we report the Pearson r aggregated over all the nine test folds combined.

\begin{algorithm}[!h]

\caption{Actor-critic Adaptive Method}
\label{alg:AC_implement}
\small
\begin{algorithmic}[1]
\item[\textbf{Notation:}] $N$: Number of folds, $J$: Number of questions
\item[\hspace{1em}]\texttt{me}:  Measure split; \texttt{err}:  Error split; \texttt{te}: Test split;
\item[\hspace{1em}]M$_{\texttt{me}}^{\texttt{J'}}$: Measure model-- trained on responses to a set of items J', predicts the ``true'' score.
\item[\hspace{1em}]E$_{\texttt{err}}^{\texttt{J'}}$[k]: Error model-- trained on responses to a set of items J', predicts error when k is the item to be added.
\item[\hspace{1em}]D$_{\texttt{i}}^{\texttt{j}}$: data split i; responses to item j; S: items administered
\item[\hspace{1em}]Y$_{\texttt{i}}$: Measure values (e.g. PHQ-9) for data split i.


\vspace{1mm}
\hrule
\vspace{1mm}
\noindent \textbf{\textsc{Training}}   
\vspace{1mm}
\hrule
\vspace{1mm}
\Function{MeasureModeling}{D$_{\texttt{me}}$}
 \For{$\texttt{j}=1,...,J$} 
\State M$^{\texttt{j}} \leftarrow $ FitRegression( D$_{\texttt{me}}$, Y$_{\texttt{me}})$
\State $\Hat Y_{\texttt{me}} \leftarrow $ M$^{\texttt{j}}$.predict(D$_{\texttt{me}})$
\State$ \Hat Y^{\texttt{j}}_{\texttt{tr}}  \leftarrow $ M$^{\texttt{j}}$.predict(D$_{\texttt{tr}})$
; $\Hat Y^{\texttt{j}}_{\texttt{te}}  \leftarrow 
 $  M$^{\texttt{j}}$.predict(D$_{\texttt{te}})$
 \EndFor
   \For{J'$ \in $ {powerSet({1,2 ... J})}} 
   \State M$_{\texttt{me}}^{\texttt{J'}}\leftarrow $ FitRegression( D$_{\texttt{me}}$, Y$_{\texttt{me}}$)
\EndFor
\EndFunction
   \Function{ErrorModeling} {D$_{\texttt{err}}$}
   \For{J$_{\texttt{temp}} \in $ {powerSet({1,2 ... J})}} 
   \For{(i $\in$ J - J$_{\texttt{temp}}$)} \# for each item not in J$_{\texttt{temp}}$
   \State J' $\leftarrow$ J$_{\texttt{temp}}$+\{i\}
   \State $ \hat{\delta}_{\texttt{err}} \leftarrow $ MeanSquaredError(Y$_{\texttt{me}}$, M$_{\texttt{me}}^{\texttt{J'}}$.predict(D$_{\texttt{err}}$))
   \State E$_{\texttt{err}}^{\texttt{J}_{\texttt{temp}}}$ [i] $ \leftarrow $ FitRegression(D$_{\texttt{err}}^{\texttt{J}_{\texttt{temp}}}$, $\hat{\delta}_{\texttt{err}}$ )
   \EndFor
   \EndFor
   \EndFunction

\vspace{1mm}
\hrule
\vspace{1mm}
\noindent \textsc{\textbf{Adaptive Testing}}
\vspace{1mm}
\hrule
\vspace{1mm}
\Function{GetNextQuestion}{itemsLeft, S}
\Return j | min. E$^{\texttt{S}}_{\texttt{err}}$[j].predict(D$^{\texttt{S}}_{\texttt{err}}$) $ \forall \texttt{j}\in$\{itemsLeft\}

\EndFunction

\For {each set of user responses d$_{\texttt{te}}$ in D$_{\texttt{te}}$:}
\State $\theta$ = $\theta_{0}\leftarrow 0$; itemsLeft $\leftarrow$ \{1... J\}; S $\leftarrow$ \{\}
\While{itemsLeft} 
\State j $\leftarrow$ \textsc{GetNextQuestion}(itemsLeft, S)
\State $\theta$ $\leftarrow $ M$_{\texttt{me}}^{\texttt{S+\{j\}}}$.predict( d$^{\texttt{j}}_{\texttt{te}}$ ); 
\State itemsLeft $\leftarrow$ itemsLeft - \{j\}; S $\leftarrow$ S + \{j\};
\EndWhile
\EndFor 
\end{algorithmic}
\end{algorithm}

\subsection{\textit{Actor-Critic} model}
\label{subsubsec:Actor-Critic}

 Based on \textit{Actor-Critic} framework used in the field of reinforcement learning \cite{grondman2012survey}, 
we design a two-model system, where the first model (Measure Model) is guided by the second model (Error Model) to take the next step adaptively. Algorithm \ref{alg:AC_implement} provides a walk-through for this model. In our case, the Measure Model learns to predict the psychometric measures directly from the all the items administered so far, whereas the Error Model learns the error (MSE) of the Measure Model over each of the unadministered items. The Error Model dictates which item to select next based on the minimum predicted error. Unlike \texttt{ALIRT}, the prediction at each step does not depend on the previous step. 
The input to \textit{Actor-critic} is predictions of the multiple ridge regression models for each question -- a continuous value as opposed to the polytomized value in \texttt{ALIRT}. 

We run a 9-fold cross validation with the same dataset split as \texttt{ALIRT} for comparability, such that D$_{\texttt{err}}$ = D$_{\texttt{poly}}$, D$_{\texttt{tr}}$ = D$_{\texttt{me}}$, and the test split D$_{\texttt{te}}$ being the same across experiments.



 \begin{table*}[ht]
 
\centering
\small
\begin{tabular}{l|ccccc||ccccc|c}

\toprule
 &
  \multicolumn{5}{c}{\textbf{Evaluated Against CTT}} &
  \multicolumn{5}{c|}{\textbf{Evaluated Against  $\hat{L}_{all}$}} &
   \\
 \textbf{Model} &
  \multicolumn{5}{c}{\textbf{Num items}} &
  \multicolumn{5}{c|}{\textbf{Num items}} &
  \textbf{Num} \\
  
\textbf{} &
  \textbf{1} &
  \textbf{2} &
  \textbf{3} &
  \textbf{4} &
  \textbf{5} &
  \textbf{1} &
  \textbf{2} &
  \textbf{3} &
  \textbf{4} &
  \textbf{5} &
  \textbf{params} \\
  \hline
RandomOrder-$\hat{L}$ & 
  0.526 &
  0.633 &
  0.669 &
  0.703 &
  0.719 &
  0.676 &
  0.806 &
  0.852 &
  0.908 &
  0.930 &
  88 \\
RandomOrder-$\widehat{Y}$ &
  0.543 &
  0.640 &
  0.675 &
  0.716 &
  0.733 &
  0.690 &
  0.813 &
  0.846 &
  0.906 &
  0.915 &
  11 \\
FixedBack-$\widehat{Y}
$ &
  0.600 &
  0.669 &
  0.701 &
0.738 &
0.749 &
  0.787 &
  0.877 &
  0.908 &
0.932 &
  0.945 &
  11\\
FixedFor-$\widehat{Y}$ &
  0.599 &
0.690 &
0.709&
  0.731 &
  0.746 &
  0.785 &
  0.880 &
  0.911 &
 0.932 &
 0.949 &
  11 \\\midrule
DecisionTree-$\hat{L} \dagger{}$ &
  0.604 &
  0.658 &
  0.679 &
  0.707 &
  0.722 &
  0.760 &
  0.831 &
  0.890 &
  0.917 &
  0.934 &
  479 \\
DecisionTree-$\widehat{Y}$ &
  0.621 &
  0.685 &
  0.717 &
  \textbf{0.740} &
  0.748 &
  0.774 &
  0.837 &
  0.897 &
  0.915 &
  0.924 &
  391 \\
ActorCritic-$\hat{L} \dagger$ &
  0.585 &
  0.656 &
  0.668 &
  0.699 &
  0.719 &
  0.748 &
  0.828 &
  0.881 &
  0.908 &
  0.929 &
  11,341 \\
ActorCritic-$\widehat{Y}$ &
  0.619 &
  \textbf{0.693} &
  0.714 &
  0.739 &
  \textbf{0.752} &
  0.765 &
  0.841 &
  0.893 &
  0.914 &
  0.926 &
  11,253 \\
\texttt{ALIRT}-$\hat{L}$&
  0.612 &
  0.669 &
  0.707 &
  0.723 &
  0.731 &
  0.816 &
  \textbf{0.897} &
  \textbf{0.935} &
  \textbf{0.955} &
  \textbf{0.965} &
  88 \\ 
\texttt{ALIRT}-$\widehat{Y} \dagger{}$ &
  \textbf{0.630} &
  0.685 &
  \textbf{0.726} &
  0.739 &
  0.746 &
  \textbf{0.828} &
  0.895 &
  0.929 &
  0.949 &
  0.955 &
  88 \\ 

  \bottomrule
\end{tabular}
\caption{Performance at depression severity assessment across ordering and scoring strategies (Pearson r). We find adaptive testing to be better than fixed ordering, and considering parameter explosion, \texttt{ALIRT} is better. 
Methods suffixed by $\hat{L}$ utilize IRT for scoring (i.e. the latent variable), while those suffixed by $\widehat{Y}$ utilize a direct estimate for scoring ($\widehat{Y} = mean(\hat{y})$ for all $y$ across administered questions). We find that the measures are consistent across both approaches. $\hat{L}_{all}$ refers to the latent score when all the 11 items are used. This means that administering just 3 items in the questionnaire based on \texttt{ALIRT} can achieve $>0.9$ correlation (Pearson r) with the latent score from using all the 11 items in the questionnaire.   $\dagger$: Significant reduction in error ($p < .05$) across multiple tests, compared to the best baseline (FixedFor-$\widehat{Y}$). The p-values for all the correlations is < 0.001. 
}
\label{tab:adaptive_strategies_depression}
\end{table*}

\subsection{Baseline Models.}
 We experiment with different ordering strategies to compare the commonly used adaptive IRT criterion Maximum Fisher Information with traditional, baseline permutations of ordering. 
 In particular, we explore three fixed-order approaches: \textbf{Random} -- 
We use a random ordering of the questions for each participant, with any of the unasked questions having an equal probability of being asked next; \textbf{Forward Selection (fixedFor)} -- 
As a fixed ordering baseline, we use forward selection to determine ordering, greedily picking the questions with the highest Pearson correlation for their polytomized item responses with the ``true'' scores; 
\textbf{Backward Elimination (fixedBack)} -- As another fixed ordering baseline, we use backward elimination based on  eliminating items with the lowest Pearson correlations of responses with the ``true'' scores.

As a more sophisticated, adaptive baseline, we also explore the \textbf{Decision Tree}, which can be seen as defining an adaptive strategy where the next best question (``feature'' in a decision tree) is picked based on the condition encountered at the current node. 
A decision tree is similar to IRT in that it can select the next best question contingent on previous responses. 
They are different in that the best splits are pre-calculated, and the next question is picked based on responses (``feature values'') at a node, whereas, for IRT, maximum Fisher information of the item parameters over all the remaining questions decides the next best question.

\subsection{Scoring Paradigms} We also compare across two scoring paradigms across all the experiments (Tables \ref{tab:adaptive_strategies_depression}, \ref{tab:gad-results}, \ref{tab:scoring-strategies}).
\textbf{Latent estimate ($\hat{L}$)} is the latent variable produced by the Item Response Theory (IRT) model. As the best latent estimate for depression (or anxiety), we consider the \textit{most informative} latent estimate to derive from \textit{all} the questions:  $\hat{L}_{all}$) to evaluate the rest of the methods against. \textbf{Classical Test Score ($\widehat{Y}$)}, on the other hand, is the average of item scores, much like scores derived from a traditional questionnaire for mental health assessment, based on Classical Test Theory (CTT). In this work, we use the PHQ-9 (GAD-7) for depression (anxiety) severity as the CTT-based score to evaluate against.

\begin{table}[!htb]
\small
\centering
\begin{tabular}{p{0.26\columnwidth}p{0.16\columnwidth}
p{0.1\linewidth}p{0.1\linewidth}p{0.1\linewidth}}
\toprule
\textbf{Model} &
\textbf{Outcome} &
\multicolumn{1}{c}{\textbf{1}  }   &
\multicolumn{1}{c}{\textbf{2} }    &
\multicolumn{1}{c}{\textbf{4} }    
\\
\midrule
RandOrder-$\widehat{Y}$ & CTT&
0.491&
0.636&
0.675\\
FixedFor-$\widehat{Y}$ &
CTT&
0.598&
0.638&
0.694\\
DecTree-$\widehat{Y}$ & 
CTT&
0.581&
0.643&
0.664\\
ActorCritic-$\hat{L}$   &
CTT &
0.561& 
0.631& 
0.672\\
ActorCritic-$\widehat{Y}$  &
CTT&
0.587&
0.658&
0.705\\
\texttt{ALIRT}-$\hat{L}$ & 
CTT &
0.600          &
0.653 & 
0.694 \\
\texttt{ALIRT}-$\widehat{Y}$              & 
CTT         & 
\textbf{0.608} &
\textbf{0.663} &
\textbf{0.707} \\ 
\midrule
RandOrder-$\widehat{Y}$ &

$\hat{L}_{all}$ &
0.603&
0.770&
0.902\\
FixedFor-$\widehat{Y}$ &
$\hat{L}_{all}$ &
0.805& 
0.877&
0.935
\\
DecTree-$\widehat{Y}$ &
$\hat{L}_{all}$&
0.760&
0.827&
0.844
\\
ActorCritic-$\hat{L}$    &
$\hat{L}_{all}$&
0.740&
0.841&
0.906\\
ActorCritic-$\widehat{Y}$&
$\hat{L}_{all}$&
0.758&
0.865&
0.929\\
\texttt{ALIRT}-$\hat{L}$             &
$\hat{L}_{all}$                  &
0.812          &
\textbf{0.904} &
\textbf{0.958} \\
\texttt{ALIRT}-$\widehat{Y}$ &
$\hat{L}_{all}$        &
\textbf{0.818} &
0.901 & 
0.952 \\
\bottomrule
\end{tabular}
\caption{Results for applying \texttt{ALIRT} for \textbf{anxiety severity}, as measured by GAD-7 as CTT- and $\hat{L_{all}}$-based scores. The reported values are Pearson r correlations with all the p-values < 0.001. 
}
\label{tab:gad-results}
\end{table}

\begin{table*}[]
\small
\centering

\begin{tabular}{p{0.59\linewidth} p{0.23\linewidth}p{0.13\linewidth}}
\toprule
\multicolumn{1}{l}{\textbf{Open-Ended Questions}} &\textbf{Shorthand} & \textbf{Word-Response Correlation with  the PHQ-9 } \\
\midrule
Describe how you generally felt the last 2 weeks, that is, how you felt on average. & Describe Mental Health & \multicolumn{1}{c}{0.61} 
\\
 Describe how you have been feeling about yourself over the last 2 weeks. & Describe Yourself & \multicolumn{1}{c}{0.61} 
\\
Over the last 2 weeks, have you been depressed or not?  & Describe Depression or Not & \multicolumn{1}{c}{0.58}
\\
Over the last 2 weeks, have you been worried or not? & Describe Worry or Not & \multicolumn{1}{c}{0.41} 
\\
Overall in your life, are you in harmony or not?  & Describe Harmony or Not & \multicolumn{1}{c}{0.54} 
\\
Overall in your life, are you satisfied or not?  & Describe Satisfaction or Not & \multicolumn{1}{c}{0.57} 
\\
Describe the nature of your physical movements over the last 2 weeks (have you for example been moving and speaking slowly; or the opposite, been fidgety and restless). & Describe Movement & \multicolumn{1}{c}{0.45} 
\\
Describe your sleep over the last 2 weeks. & Describe Sleep & \multicolumn{1}{c}{0.44} 
\\
Describe your concentration over the last 2 weeks. & Describe Concentration & \multicolumn{1}{c}{0.44} 
\\
Describe your appetite for food over the last 2 weeks.  & Describe Appetite  &\multicolumn{1}{c}{0.36} 
\\
Describe your energy level over the last 2 weeks.  & Describe Energy & \multicolumn{1}{c}{0.51} 
\\
\bottomrule
\end{tabular}
\caption{The questions administered to participants in our dataset, along with their shorthand used in this paper. Pearson r is reported for each of the question's word response scores with the self-reported PHQ-9 scores.
}
\label{tab:prompts}
\end{table*}

\noindent Across all the experiments described, the folds are kept consistent (including baselines).

\section{Dataset}
\label{sec:dataset}

\begin{figure}[!ht]
    \centering
    \includegraphics[width=8.cm]
{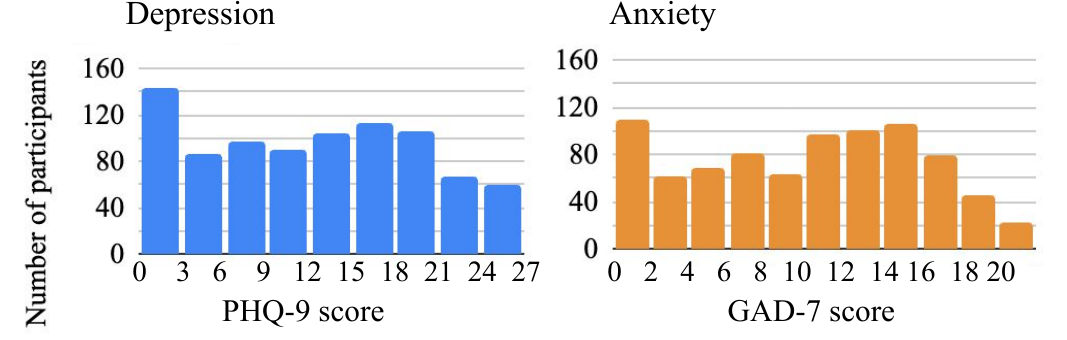}
\caption{Distribution of depression and anxiety scores of participants in the dataset described in \S\ref{sec:dataset}.}
\label{fig:participant_dist}
\end{figure}

Our dataset consists of open-ended language answers to eleven questions and two self-diagnostic
tests in the form of closed-ended rating scales. 
Participants were recruited online from Mechanical Turk (N = 528; 2018-05-05) and Prolific (N = 419; 2018-11-28), 
where they were paid \$3 and £3, respectively to participate. 
The MT data set included attention checks (e.g., ``On this item, answer alternative 3''), which 64 participants failed and thus were removed. The Prolific study included a screening procedure where 260 participants had reported being diagnosed with Major Depressive Disorder and/or Generalised Anxiety Disorder before being invited to participate; 159 participants were not screened.
The open-ended questions, shown in Table \ref{tab:prompts}, concerned mental health and well-being, including: 1. General mental health, 2. Depression, 3. Anxiety, 4. Harmony, 5. Satisfaction; and some mental health-related symptoms: 6. Movement, 7. Sleep, 8. Concentration, 9. Appetite, 10. Energy, and 11. Self-perception. 
The participants were asked to respond using at least five descriptive words for the mental health questions (1-3), three descriptive words for the well-being questions (4-5), and two descriptive words for the symptom questions (6-11). The mean PHQ-9 score across the cohort was 11.98 with a standard deviation of 7.76, and the mean GAD-7 was 10.16 with a standard deviation of 6.23. The distribution of participants across PHQ-9 and GAD-7 scores are given in Figure \ref{fig:participant_dist}. 
The validated scale for depression was the Patient Health Questionnaire 9-item aka the PHQ-9 \cite{kroenke2001phq}, and for anxiety, the Generalised Anxiety Disorder 7-item scale aka the GAD-7 \cite{spitzer2006brief}.


We model a single latent score in this dataset, as opposed to modeling multiple mental health conditions simultaneously with multidimensional IRT models \cite{chalmers2012mirt}. This choice is justified in Appendix \ref{appendix:dimensionality}.

\section{Results \& Discussion}

\noindent We report the performances of the various adaptive strategies, in comparison with the baselines and across scoring methods, in Table \ref{tab:adaptive_strategies_depression}.




\subsection{Adaptive Strategies}

Table \ref{tab:adaptive_strategies_depression}  examines the difference between two scoring methods with experiments run for depression severity assessment. For CTT-based scoring, we use a simple averaging of predicted measures over the selected questions, which is limited by the accuracies of the $11$ individual question models ($\widehat{Y}$). $\hat{L}$ is the latent estimate produced by the IRT model. Each of these is evaluated against a ``true'' CTT score (PHQ-9) and a ``true'' latent score which is the latent estimate obtained by simultaneous parameter estimation with all the questions ($\hat{L}_{all}$).
We find that adaptive strategies tend to perform better than the baselines. Among the three adaptive strategies used to directly predict the psychometric measures in Table \ref{tab:adaptive_strategies_depression}, 
we find that the \texttt{ALIRT}-$\widehat{Y}$ performs best when compared against the CTT score,
and \texttt{ALIRT}-$\hat{L}$ performs best when compared to $\hat{L}_{all}$. The differences in correlations become less evident as the number of items administered increases due to the convergence of items picked across different strategies. 

The \textit{Actor-Critic} model has $2^{N} - 1$ score prediction models and $N.(2^{N-1} - 1)$ error prediction models (see Appendix \ref{appendix:compAC}) trained on each combination of questions to pick out the best question to administer. Despite this, the performance boost that could be afforded by the computational complexity is not always significant, and \texttt{ALIRT} performs similarly (or better) despite a much smaller number of parameters and shorter runtime. It is notable that \texttt{ALIRT}, which uses Maximum Fisher Information for adaptive ordering, does \textit{not} try to optimize for errors/correlation with the ``true'' scores, but the ordering produced by it largely helps across both the scoring paradigms, which demonstrates the utility of IRT in being able to capture inherent associations without direct supervision. 

These findings are fairly consistent for anxiety severity assessment as well, evaluated against GAD-7, as seen in Table \ref{tab:gad-results}. This indicates that adaptive language-based assessment could be extended to other common, standardized assessments as well.

\begin{table}[]
\centering
\small
\begin{tabular}{p{0.2\linewidth}p{0.07\linewidth}r|p{0.07\linewidth}r||r}
\toprule
\multicolumn{1}{r}{Eval against:}&\multicolumn{2}{c}{CTT}     &\multicolumn{2}{c}{$\hat{L}_{all}$}           &           \multicolumn{1}{c}{\textbf{Num} }                                          \\
\textbf{Approach} &
  \multicolumn{1}{c}{\textbf{2}}   &
  \multicolumn{1}{c}{\textbf{4}} &
  \multicolumn{1}{c}{\textbf{2}}  &
  \multicolumn{1}{c}{\textbf{4}}  &  \textbf{Params}\\
  \midrule
\textit{Actor-Critic}\\
\midrule
$\widehat{Y}$ &
  0.693 &
  0.739 &
  0.841 &
  0.914 &
  11,253\\

Regr ($\hat{Y}$) &
  \textbf{0.694} &
  \textbf{0.741} &
 0.873&
  0.927 &
  11,253
  \\
Regr (X)                    &
  0.693 & 
  0.734          &
  0.857 &
  0.922&
  112,530\\

  \midrule
\texttt{ALIRT}\\
\midrule
$\hat{L}$ &
  0.669 &
  0.723 &
  \textbf{0.897} &
  \textbf{0.955} &
  88\\
Regr ($\hat{Y}$) &
  0.685 &
  0.736 &
  0.896 &
  0.947 &
  11,341\\
Regr (X)                    & 
0.685          & 
0.740 &
  0.883 &
  0.934 &
  112,618\\ 
\bottomrule
\end{tabular}
\caption{Performance of depression severity assessment across ordering strategies for regression-based scoring strategies. We compare using regression on the item scores and on the word embeddings to the best from Table \ref{tab:adaptive_strategies_depression}-- $\widehat{Y}$ and $\hat{L}$. The reported values are Pearson r, with p-values < 0.001.
}
\label{tab:scoring-strategies}
\end{table}

\subsection{Scoring strategies}
We note from table \ref{tab:adaptive_strategies_depression} that there is merit to both the scoring paradigms, with CTT offering a widely accepted, standardized, fixed scale with supervision in every step, whereas IRT allows semi-supervision and can adapt the scale according to the response behavior of the cohort of participants. 
We compare the two scoring strategies to regression-based scoring as well, where instead of averaging the scores over the selected questions, we use regression to train prediction models to output a score, with the item response as input. Table \ref{tab:scoring-strategies} compares the various scoring strategies and how they correlate to CTT-based ``true" scores and IRT-based ``most informative'' scores. 

\paragraph{Regression over word embeddings -- Regr(X)}
The input is the item response word embeddings. We find that this method does not really fare better across both \textit{Actor-Critic} and \texttt{ALIRT}. Since we use 10 dimensional word embeddings, the number of parameters is increased tenfold, which could cause the model to overfit. Moreover, the method is unrealistic when scaled up to more questions due to parameter explosion.

\paragraph{Regression over predicted scores -- Regr($\hat{Y}$)}
Item response scores are used as input to the model, and thus we can re-use all the models trained for \textit{Actor-Critic} approach. While there is still risk of parameter explosion if there were more questions, the method does not demand more compute and seems to improve the correlations of the predicted scores in the \textit{Actor-Critic} setting.

\begin{figure}[!ht]
    \centering
    \includegraphics[width=8cm]{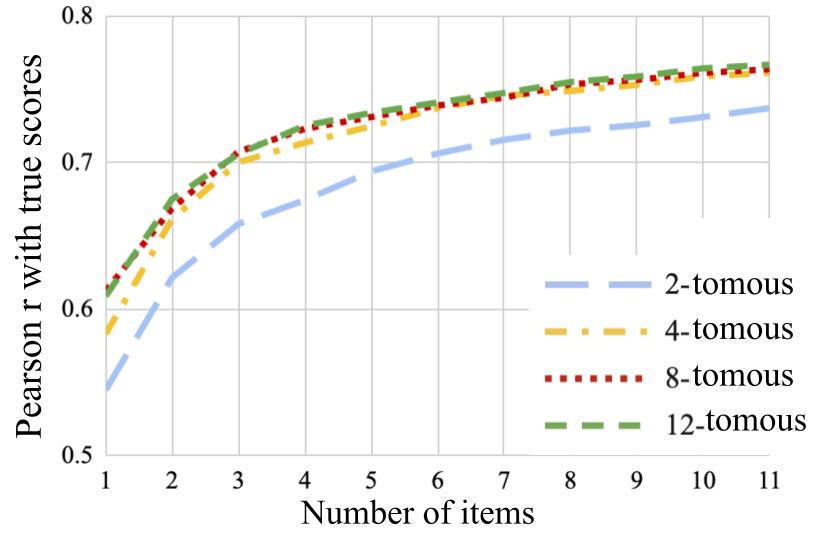}
    \caption{The correlation of the latent scores with the ``true" (PHQ-9) scores for various polytomization levels across the number of items. 12-tomous model is likely to be overfit and does not offer significant advantage over our initial choice of 8. 
    }
\label{fig: discretization-level}

\end{figure}

\subsection{Optimal Discretization}

 Some information can be lost when numeric values are polytomous \cite{catlett1991changing}. For the purposes of use in adaptive testing with IRT, it is unclear how discretized the values should be. On the one hand, there can be more information loss with coarse-grained discretization (i.e. less number of choices in the rating scale); and on the other hand, fine-grained discretized (i.e too many choices in the rating scale) results in too many parameters with respect to data size.
 The result of our experiment is seen in Figure \ref{fig: discretization-level} where we experiment for 2, 4, 8 and 12. A polytomization of 8 works just as well as 12 with $\frac{2}{3}$ the number of parameters. We also found that a polytomization of 13 or above results in missing values -- resulting in ill-fit characteristic curves used in the IRT model.

\subsection{Most Informative Questions: Depression Severity}
\begin{figure}[!htbp]
    \centering
\includegraphics[width=7.5cm]{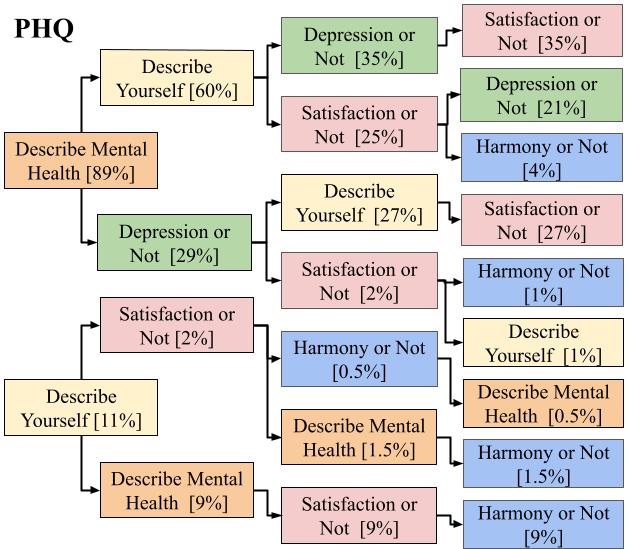}
    \caption{
    Flowchart of the items picked at $n^{th}$ question using \texttt{ALIRT}. The selections of questions for the first few items is rather sparse. Since the latent variable estimate does achieve a high correlation with the classical psychometric measures in 3-5 questions, it hints at the irrelevance of some questions towards the psychometric measure despite high individual feature correlations. 
}
    \label{fig: heatmap-alirt}

\end{figure}

Based on Table \ref{tab:adaptive_strategies_depression}, we find that \texttt{ALIRT} achieves a high correlation ($r > 0.7$) to standardized assessments with 3 questions. Figure \ref{fig: heatmap-alirt} tells us that the questions are not highly personalized-- for the first 4 items, only 6 out of total 11 questions are administered, with general mental health questions (``Describe Yourself" and ``Describe Mental Health") being the most informative first questions to ask. None of the symptom questions are asked at all, possibly hinting at the redundancy of such questions in language-based assessments for depression severity.

\section{Related Work}
Over the past decade, researchers have been exploring 
techniques for mental health assessment~\cite{coppersmith2015clpsych}. Initial studies inspired by leveraging communication in social media, indicated that NLP models could moderately accurately predict self-disclosed mental health conditions or events~\cite{coppersmith2015clpsych,de2016discovering}, scores from self-report mental health questionnaires~\cite{schwartz2014towards,chancellor2020methods}, and achieve scores aligned with standard screening surveys when compared to clinical records of depression~\cite{eichstaedt2018facebook}.
However, such methods only work well with a fairly active social media usage~\cite{kern2016gaining}. 
While some have proposed methods to utilize transformers with smaller datasets~\cite{ganesan2021empirical}, such an approach is still limited to those willing to share such data or having any of it at all. 
Further, it has recently been shown that the accuracy of language-based assessments can reach even greater ($r > 0.8$) when the assessment is based on prompting participants for language responses related to mental health, mirroring standard questionnaires but using language responses instead \cite{kjell2022natural}. 
Still, past work has mostly been validated against summed, or averaged, questionnaire scores, while here we consider improved measurement paradigms that rely on latent variables, such as item-response theory~\cite{reise2009item}. 

Within the domains of NLP, IRT has been used in chatbot evaluation~\cite{sedoc-ungar-2020-item}, for textual entailment \cite{lalor2016building}. 
 IRT has also been used to impute missing data \cite{pliakos2019integrating} and to compare different ML classifiers at an instance level. 
Feature/question selection (an NP-complete problem) has also been explored with IRT over a number of fixed selection, ranking, and ordering methods in the recent years \cite{abdel2009constructing,kline2020novel,coban2022new}. 
In a related study, \cite{coban2022irtext} applied IRT to linguistic data, converting language into a term-document matrix for feature selection. However, our approach in adaptive language-based assessment extends beyond fixed feature selection settings. We aim to dynamically adapt to each data sample, facilitated by IRT-based ordering. It's important to note that adaptive language testing differs from personalized recommendation systems. While the latter emphasizes item similarity, language-based testing strives to precisely assess users' latent traits, setting it apart from recommendation systems designed for preferences.

\section{Conclusion}

Mental health issues vary widely across individuals, suggesting the need for assessments that can enable wide ranging symptoms and be adaptive to the individual. 
We introduced the task of adaptive language-based assessment for eliciting the most informative responses as well as developed and explored two methods to perform the task, \texttt{ALIRT} and the \textit{Actor-Critic} methods, along with a suite of more straight-forward approaches.  
Evaluated against depression severity scores derived from 11 questions, \texttt{ALIRT} was able to capture over 90\% of the variance explained ($R^2$) after only 4 questions while optimal fixed ordering approaches needed at least 7, suggesting patient time could be saved with this approach.  
We further saw that a regression approach that tries to optimally weight question-scores had only minor benefits over the IRT-based ($\hat{L}$), that \texttt{ALIRT} generalized to assessing anxiety in addition to depression, and that symptom-focused questions were not as informative (never chosen early) as compared to broader questions. 
The adaptive approach, in general, can significantly reduce the number of questions required to achieve high validity, as well as yield insights into the questions that produce the most informative responses suggesting better question/prompt creation. 

\section{Acknowledgements}
We would like to thank the reviewers for their valuable feedback that helped us improving this paper.

This work was supported in part by a grant from the CDC/NIOSH (U01 OH012476), a grant from the NIH-NIAAA (R01 AA028032) and a DARPA Young Faculty Award grant \#W911NF-20-1-0306 awarded to H. Andrew Schwartz at Stony Brook University.
The conclusions contained herein are those of the
authors and should not be interpreted as necessarily representing the official policies, either expressed or implied, of DARPA, NIH, any other
government organization, or the U.S. Government.

\section{Limitations}

This work has a few key limitations: for Classical Test Theory (CTT), we assessed outcomes using self-report questionnaires, specifically PHQ-9 and GAD-7. However, relying on self-reporting in surveys may not ensure complete reliability for diagnostic accuracy. Nevertheless, such self-reported measures have demonstrated consistent associations with diagnoses, proving valuable in clinical assessment and treatment contexts beyond diagnosis \cite{kroenke2001phq}. For instance, anxiety scores from self-reported surveys have shown strong correlations with significant real-world outcomes like mortality \citep{kikkenborg2014anxiety}. To validate the assessments proposed in this study, it is crucial to evaluate them against clinical outcomes.

The study was limited by the number of data points and use of descriptive words in English, instead of open-ended texts, due to which we use word embeddings instead of contextual embeddings. While the results in this paper that make the case for adaptive testing should likely translate to other domains including open-ended questions and response domains, we leave that direction open for future work. Instead, we view our work as a first step in integrating adaptive testing into chatbot-style mental health assessments, with a small dataset of descriptive word responses.

\section{Ethical Considerations}

The dataset used was collected from participants in Prolific and Amazon Mechanical Turk, who were paid to respond to the 11 descriptive questions, along with PHQ-9 and GAD-7 questionnaires. The participants were English-speaking and geographically located in the UK. All of the data is anonymized. 
The research was approved by an academic institutional ethics review board (exempt status).

This method could potentially be used in the wild-- social media posts disclosing diagnoses could be abused to train larger models and track people's latent psychological traits at each utterance in their language, exposing vulnerable people on social media to potential exploitation. 

However, as NLP advances in enhancing human-focused applications, such as improving mental health assessment, the balance between considerations for human privacy and open data sharing becomes crucial. In this instance, the data used was shared only with consent for academic research, and open sharing violates trust with participants and ethical review board agreements. \citet{benton2017ethical} extensively discusses these issues. While the ideal is to release everything while preserving privacy, the limited availability of data suggests an imperative for those with access to share our work openly within ethical guidelines.




\bibliographystyle{acl_natbib}
\bibliography{anthology,custom}
\setcounter{table}{0}
\renewcommand{\thetable}{A\arabic{table}}
\appendix

\section{Dimension Reduction on Contextual Embeddings}
\label{appendix: roberta_large}

Our experiment was limited to static embeddings trained specifically on the mental health domain, which was due to the data scarcity and format (descriptive words), leading to the need for a small embedding size. For comparability and to show the utility of adaptive testing with other embeddings, we experiment with dimension reduction on RoBERTA-large model. We separately collect sets of five words describing daily emotions, mood, feelings etc. from 572 users, for about 30 days each. RoBERTA-large (1024 dim) embeddings are extracted for each of these sets, on which PCA is applied to reduce the dimensions from 1024 to 10. The reduction was then applied to all the questions in Table \ref{tab:prompts}, and then trained \texttt{ALIRT} and \textit{Actor-Critic} models. The results are reported in Table \ref{tab:roberta_large10_results}. 
The same procedure was repeated with GloVe embeddings (cased, trained on Common Crawl) by learning a reduction on 300 dimensional GloVe vectors to 10 dimensions for fair comparison. The results for GloVe are reported in \ref{tab:glove_10_results}.

\begin{table}[!h]
\centering
\small
\centering

\begin{tabular}{l |c |r r r}
\toprule
 Method & Eval. against & 1 & 2 & 4 \\
 \midrule
FixedFor-$\widehat{L}$ & CTT & 0.582& 0.653& 0.708\\
FixedFor-$\widehat{Y}$ & CTT & 0.591& 0.670& \textbf{0.718}\\
ActorCritic-$\hat{L}$ & CTT & 0.575& 0.652& 0.682\\
ActorCritic-$\widehat{Y}$ & CTT & \textbf{0.604}& \textbf{0.678}& 0.705\\
\texttt{ALIRT}-$\hat{L}$ & CTT & 0.592& 0.651& 0.706\\
\texttt{ALIRT}-$\widehat{Y}$ & CTT & 0.596& 0.661& 0.715\\
\midrule
FixedFor-$\hat{L}$ & $\hat{L}_{all}$ & 0.762& 0.857& 0.934\\
FixedFor-$\widehat{Y}$ & $\hat{L}_{all}$ & 0.773& 0.864& 0.929\\
ActorCritic-$\hat{L}$ & $\hat{L}_{all}$ & 0.740& 0.828& 0.913\\
ActorCritic-$\widehat{Y}$ & $\hat{L}_{all}$ & 0.769& 0.840& 0.908\\
\texttt{ALIRT}-$\hat{L}$ & $\hat{L}_{all}$ & 0.784& 0.873& \textbf{0.944}\\
\texttt{ALIRT}-$\widehat{Y}$ & $\hat{L}_{all}$ & \textbf{0.792}& \textbf{0.878}& 0.938\\
\bottomrule

\end{tabular}

\caption{Comparison of fixed and adaptive strategies with 10-dimensional contexutual embeddings reduced from RoBERTA-large, evaluated against PHQ-9 for CTT and $\hat{L}_{all}$-- the latent score derived when using all the items -- for IRT . }
\label{tab:roberta_large10_results}
\end{table}

\texttt{ALIRT} is a better choice when using RoBERTA-large as well, especially when using IRT scoring strategy, but does not compromise much on the performance given the number of parameters in Classical Test Theory too. Forward selection is comparable to adaptive testing among fixed ordering methods. However, the difference between fixed and adaptive strategies is not as significant as when using static embeddings. This can be explained with the context-independent word responses in the dataset used, where contextual embeddings do not seem to improve the predictive power.

\begin{table}[!h]
\centering
\small
\centering

\begin{tabular}{l |c |r r r}
\toprule
 Method & Eval. against & 1 & 2 & 4 \\
 \midrule
FixedFor-$\widehat{L}$ & CTT & 0.626& 0.703& 0.732\\
FixedFor-$\widehat{Y}$ & CTT & 0.637& 0.714& 0.747\\
ActorCritic-$\hat{L}$ & CTT & 0.605& 0.695& 0.729\\
ActorCritic-$\widehat{Y}$ & CTT & 0.628& \textbf{0.723}& \textbf{0.750}\\
\texttt{ALIRT}-$\hat{L}$ & CTT & 0.630& 0.660& 0.719\\
\texttt{ALIRT}-$\widehat{Y}$ & CTT & \textbf{0.644}& 0.712& 0.748\\
\bottomrule

\end{tabular}

\caption{Comparison of fixed and adaptive strategies with 10-dimensional word embeddings that were reduced with GloVe embeddings, evaluated against PHQ-9 for CTT. Consistent with the results observed with LSA and RoBERTA-large embeddings, the adaptive methods perform better than fixed. Further, the effect observed with GloVe is comparable to that of LSA as opposed to RoBERTA-large since it is non-contextual and better suited for descriptive words rather than open-ended language.}
\label{tab:glove_10_results}
\end{table}

\section{Computational Complexity of the \textit{Actor-critic} model}
\label{appendix:compAC}
For N items, there are  $2^{N}-1$ combinations of items, and therefore, $2^{N}-1$ error prediction models.  
For $N$ total questions and k questions administered so far, the number of combinations of questions left is $\binom{N}{N-k} =\binom{N}{k} $. Number of items that could be picked next is ($N-k$) Adding them over all the possibilities:

\begin{equation*}
(N-1).\binom{N}{1} + ... + (N-N).\binom{N}{N} 
\end{equation*}

\begin{equation*}
= \sum_{1}^{N} (N-k).\binom{N}{k} 
= -N+\sum_{0}^{N} (N-k).\binom{N}{k} 
\end{equation*} 

\begin{equation*}
  = -N + N\sum_{0}^{N}\binom{N-1}{k}
   = N(2^{N-1}-1)
\end{equation*} 
We arrive at a complexity of $O(N.2^{N})$.

\section{Dataset Dimensionality}
\label{appendix:dimensionality}
Item response theory is a form of factor analysis \cite{takane1987relationship}. Therefore, we perform two tests to ensure the feasibility of our dataset. \cite{dziuban1974correlation} Kaiser–Meyer–Olkin (KMO) test  \cite{kaiser1974educational} checks sampling adequacy for each feature based on the correlation matrix and produces a KMO value between 0-1. The higher the KMO value is, the better suited the data is for factor analysis. Our dataset has a KMO value of 0.924, which makes it highly suitable for factor analysis.
 We also perform the Bartlett Test of Sphericity on our dataset to determine the number of significant factors. \cite{gorsuch1973using} 
The test results in a p-value $<.001$, which indicates that the IRT latent variable should indeed capture the features, with the Kaiser criterion indicating there is just 1 latent factor. 

\section{IRT parameters for \texttt{ALIRT}}
\label{appendix:irt}
The polytomous model fits a 2-parameter (2PL) characteristic curve for each polytomous threshold for each item. 2PL item characteristic curve is typically modeled with two parameters:
\begin{equation*}
    P(\theta) = \frac{1}{1+e^{-\alpha(\theta - \beta)}}
\end{equation*}
where $\beta$ is the difficulty parameter (midpoint of the slope; models how ``difficult'' an item is) and $\alpha$ is the discriminant (slope of the midpoint; it models how well an item discriminates between participants that score higher/lower than the difficulty).
For polytomous IRT modeling, if the responses are polytomized to K values [0,1, ... K-1, K], then there are K-1 logistic characteristic curves learned for each threshold: between 0 and 1, between  1 and 2 ... and between K-1 and K. In our case, a single discriminant $\alpha$ is learned across all the K-1 curves per item. Therefore, for j$^{th}$ item and k$^{th}$ curve, the item characteristic function is:
\begin{equation*}
    P(\theta^{j}_{k}) = \frac{1}{1+e^{-\alpha^{j}(\theta - \beta^{j}_{k})}} - \frac{1}{1+e^{-\alpha^{j}(\theta - \beta^{j}_{k-1})}}
\end{equation*}
The total number of parameters for J questions, with K-tomous responses is therefore J x K.
Maximum Fisher Information (MFI) is the objective used by \texttt{ALIRT} to pick the next best question. This is calculated as the derivative of log probabilities at the current latent estimate using the item characteristic functions. \cite{hald1999history} MFI picks the question with highest variance in the estimate of the score/latent variable.
The latent variable is clipped between -6 and +6.

\end{document}